\documentclass[10pt,twocolumn,letterpaper]{article}

\usepackage{iccv}              

\usepackage{multirow}
\usepackage{colortbl}
\usepackage{arydshln,hhline}

\definecolor{iccvblue}{rgb}{0.21,0.49,0.74}
\usepackage[pagebackref,breaklinks,colorlinks,allcolors=iccvblue]{hyperref}

\def\paperID{6278} 
\def\confName{ICCV}
\def\confYear{2025}

\title{SFUOD: Source-Free Unknown Object Detection}

\author{Keon-Hee Park\\
Kyung Hee University\\
Yongin, Republic of Korea\\
{\tt\small pgh2874@khu.ac.kr}
\and
Seun-An Choe\\
Kyung Hee University \\
Yongin, Republic of Korea \\
{\tt\small dragoon0905@khu.ac.kr}
\and
Gyeong-Moon Park\thanks{Corresponding author.}\\
Korea University\\
Seoul, Republic of Korea\\
{\tt\small gm-park@korea.ac.kr}
}

\begin{document}
\maketitle

\newcolumntype{g}{>{\columncolor{Gray!15}}c}
\newcolumntype{b}{>{\columncolor{RoyalBlue!10}}c}
\newcolumntype{y}{>{\columncolor{yellow!15}}c}

\begin{abstract}
Source-free object detection adapts a detector pre-trained on a source domain to an unlabeled target domain without requiring access to labeled source data. While this setting is practical as it eliminates the need for the source dataset during domain adaptation, it operates under the restrictive assumption that only pre-defined objects from the source domain exist in the target domain. This closed-set setting prevents the detector from detecting undefined objects.
To ease this assumption, we propose \textbf{S}ource-\textbf{F}ree \textbf{U}nknown \textbf{O}bject \textbf{D}etection (\textbf{SFUOD}), a novel scenario which enables the detector to not only recognize known objects but also detect undefined objects as unknown objects. 
To this end, we propose \textbf{CollaPAUL} (\textbf{Colla}borative tuning and \textbf{P}rincipal \textbf{A}xis-based \textbf{U}nknown \textbf{L}abeling), a novel framework for SFUOD. Collaborative tuning enhances knowledge adaptation by integrating target-dependent knowledge from the auxiliary encoder with source-dependent knowledge from the pre-trained detector through a cross-domain attention mechanism. Additionally, principal axes-based unknown labeling assigns pseudo-labels to unknown objects by estimating objectness via principal axes projection and confidence scores from model predictions.
The proposed CollaPAUL achieves state-of-the-art performances on SFUOD benchmarks, and extensive experiments validate its effectiveness. Our code is available at \href{https://github.com/KU-VGI/SFUOD}{SFUOD}.
\end{abstract}
\vspace{-6mm}    
\section{Introduction}
\label{sec:intro}
Domain adaptive object detection~\cite{chen2018domain,deng2021unbiased,deng2023cross,hsu2020every,li2022sigma,vs2021mega} aims to transfer knowledge from a source to a target domain and is widely studied for its practical applications. Especially, source-free object detection (SFOD)~\cite{li2021free,liu2023periodically,khanh2024dynamic,hao2024simplifying,vs2023instance,li2022source,chu2023adversarial} focuses on adapting source-trained models to the unlabeled target domain without requiring access to labeled source data. SFOD is a realistic scenario that addresses data privacy and storage constraints. However, SFOD assumes a closed-set scenario, where the source and target domains share the same set of class labels, struggling to detect objects not defined in the source domain. In real-world applications (e.g., self-driving), 
detectors must recognize unknown objects to handle unforeseen events (\eg, identifying pedestrians or animals on the road to prevent accidents), but SFOD limits the detector from detecting unknown objects.

\begin{figure}
    \centering
    \includegraphics[width=\columnwidth]{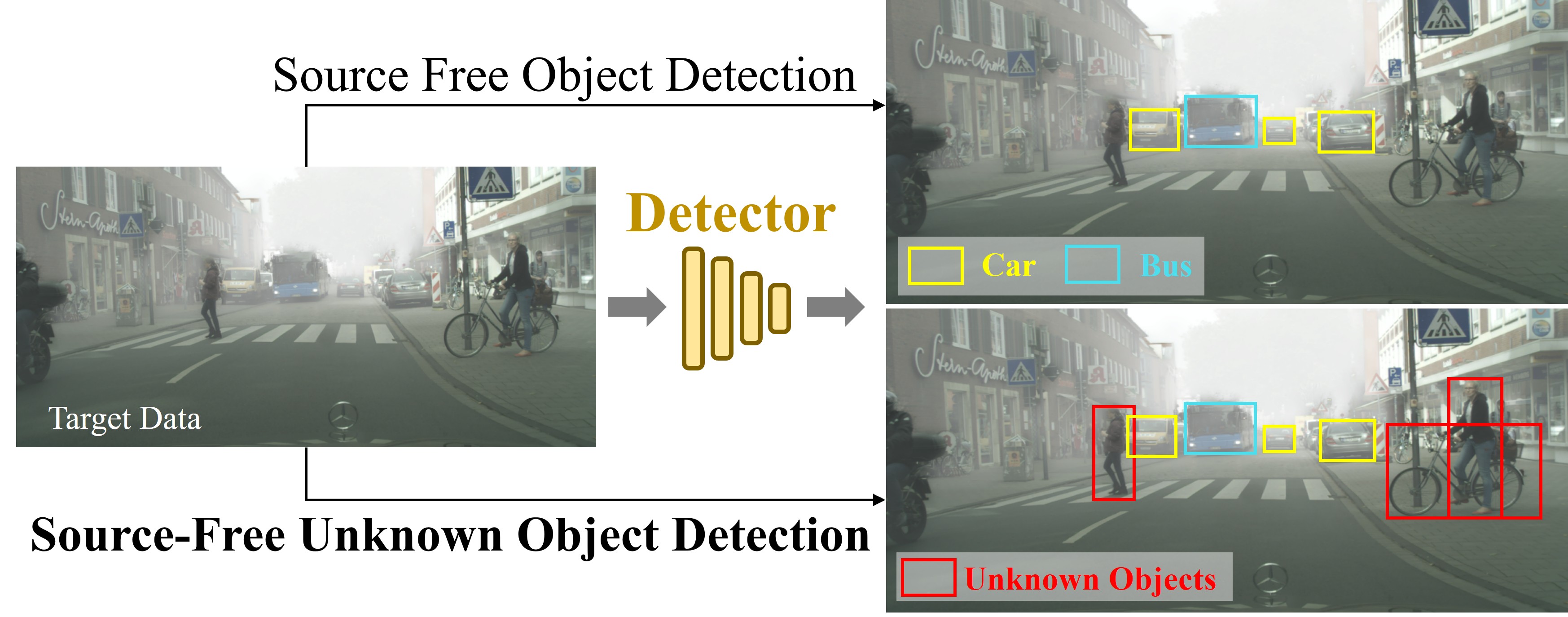}
    \caption{Description of the proposed SFUOD. Yellow and skyblue boxes denote known objects (\eg, ``Car'', ``Bus''). Red boxes denote unknown objects (\eg, ``Person'', ``Bicycle'').\vspace{-7mm}}
    \label{fig:description_sfuod}
\end{figure}

To address the limitation of detecting unknown objects in source-free object detection, we propose a novel Source-Free Unknown Object Detection (SFUOD) scenario. The SFUOD scenario requires not only adapting a source-trained detector to the target domain for known objects but also detecting undefined objects as unknown. As shown in~\Cref{fig:description_sfuod}, traditional source-free object detection adapts a detector to recognize only pre-defined objects from the source domain (\eg, ``Car'', ``Bus''). In contrast, the detector can also detect unknown objects (\eg, ``Person'', ``Bicycle'') in our new scenario. Since unknown objects frequently appear in the real world, the proposed SFUOD is a realistic and challenging scenario as it simultaneously requires knowledge transfer for known objects and detecting unknown objects.
\begin{figure}
  \centering
  \begin{subfigure}{\columnwidth}
    \includegraphics[width=\columnwidth]{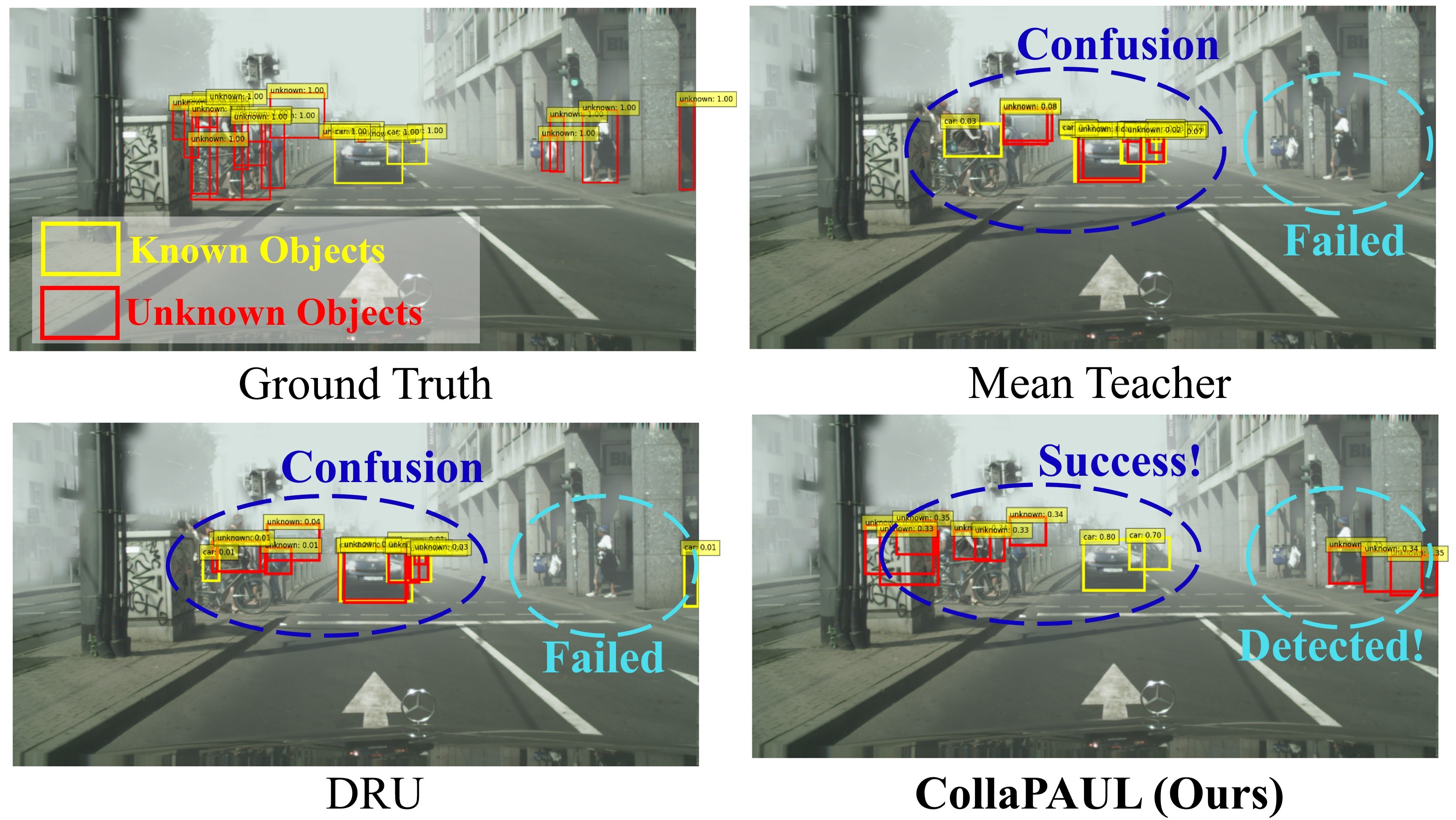}
    \caption{Qualitative comparison in the source-free unknown object detection.}
    \label{fig:motiv_a}
  \end{subfigure}
  \vfill
  \begin{subfigure}{\columnwidth}
  \centering
    \includegraphics[width=\columnwidth]{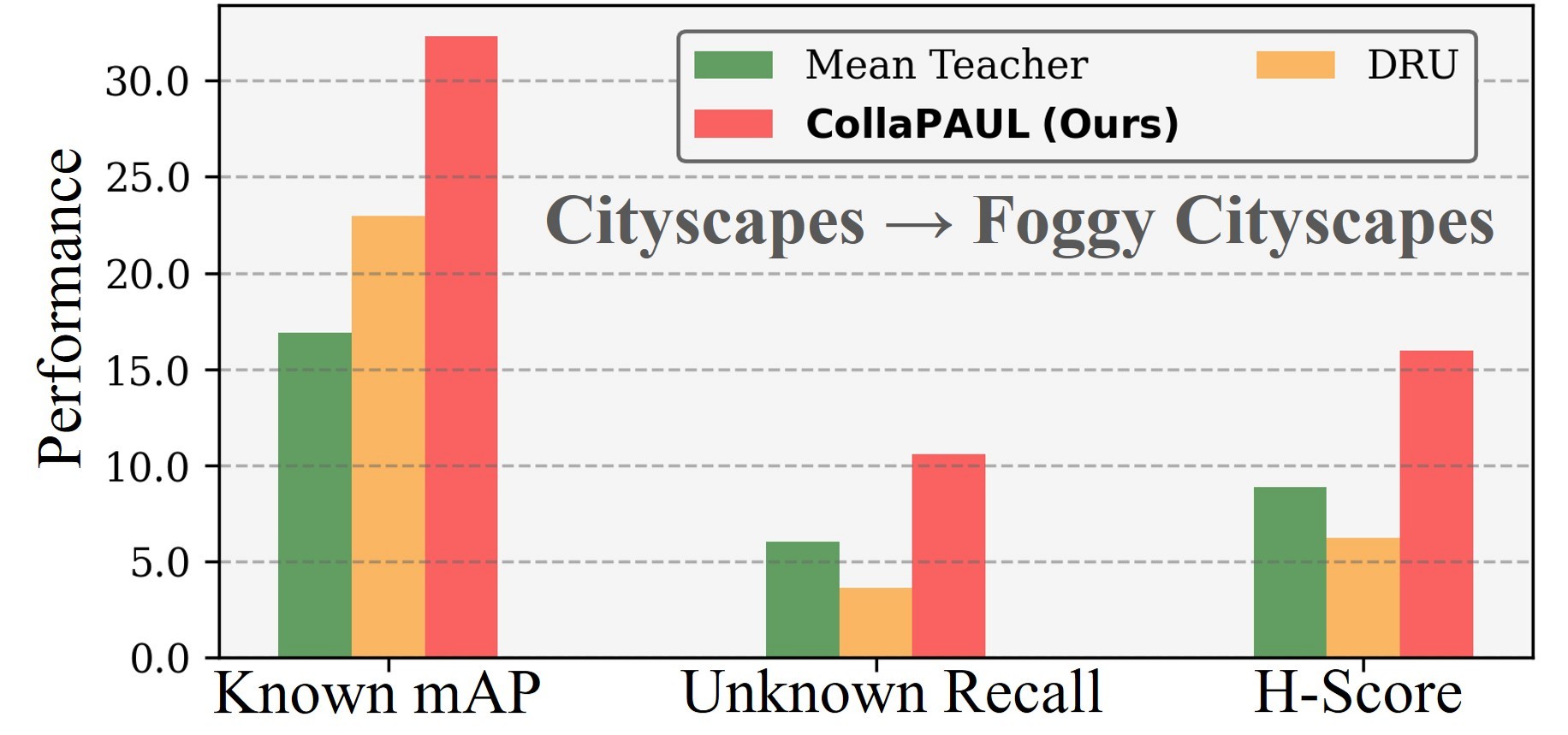}
    \caption{Quantitative comparison of Known mAP, U-Recall, and H-Score. \vspace{-2mm}}
    \label{fig:motiv_b}
  \end{subfigure}
  \caption{\Cref{fig:motiv_a} presents a qualitative comparison of our method with existing SFOD approaches in Cityscapes $\rightarrow$ Foggy Cityscapes. Prior methods often misclassify known and unknown objects (blue-colored section) and fail to detect unknown objects (sky blue-colored section). Yellow and red boxes indicate known and unknown objects, respectively. \Cref{fig:motiv_b} provides a quantitative comparison of our method with prior SFOD methods. \vspace{-7mm}}
  \label{fig:motiv}
\end{figure}

In the SFUOD scenario, existing SFOD methods struggle with both adapting to known objects and detecting unknown ones. Existing approaches like DRU~\cite{khanh2024dynamic} use the Mean Teacher (MT) framework~\cite{tarvainen2017mean}, where a teacher model generates pseudo-labels for a student model, and the teacher is updated via exponential moving average (EMA)~\cite{tarvainen2017mean} to mitigate domain shift by transferring source-dependent knowledge into the target domain.
However, since the teacher model trained in the source domain lacks knowledge of unknown objects, it cannot successfully generate pseudo-labels for unknown objects. As a result, it struggles to learn unknown objects effectively and fails to detect unknown objects. Additionally, applying source-dependent knowledge to unknown objects leads to knowledge confusion between known and unknown classes, further exacerbating domain shift and hindering adaptation to known objects.

To validate this, we tested SFOD methods (\eg, MT~\cite{tarvainen2017mean}, DRU~\cite{khanh2024dynamic}) to the proposed SFUOD scenario. The detector was trained on Cityscapes~\cite{cordts2016cityscapes} as the source domain and adapted to the unlabeled Foggy Cityscapes~\cite{sakaridis2018semantic} as the target domain. Following the SFUOD setting, the model was trained to classify only vehicles (e.g., Car, Truck, and Bus) using labeled source data, while the target domain remained unlabeled. To assign pseudo-labels to unknown objects, we employed confidence-based pseudo-labeling, a common approach in unsupervised domain adaptation~\cite{saito2018open,fu2020learning,choe2024open}.
As shown in~\Cref{fig:motiv_a}, existing SFOD methods misclassify known objects as unknown and vice versa due to knowledge confusion, often failing to detect unknown objects due to ineffective pseudo-labeling. Furthermore, \Cref{fig:motiv_b} illustrates that knowledge confusion and inaccurate unknown pseudo-labels lead to low mAP for known objects and low detection for unknown objects, respectively.
To sum up, SFUOD has two main challenges: 1) \textit{knowledge confusion} due to source-dependent knowledge hinders effective knowledge transfer, and 2) \textit{ineffective unknown pseudo-labeling} fails to detect unknown objects.

To this end, we propose \textbf{CollaPAUL} (\textbf{Colla}borative tuning and \textbf{P}rincipal \textbf{A}xis-based \textbf{U}nknown \textbf{L}abeling) for the new SFUOD scenario, employing the mean teacher approach to effectively adapt knowledge of known objects while learning unknown objects.
CollaPAUL comprises two key components. First, collaborative tuning integrates source-dependent knowledge from the student model with target-dependent knowledge from an auxiliary target encoder. To extract target-dependent knowledge, the target encoder applies truncated reconstruction via Singular Value Decomposition (SVD), and cross-domain attention is then employed to fuse source and target features, allowing the student model to learn richer representations through collaborative tuning. 
Second, Principal Axis-based Unknown Labeling (PAUL) estimates the principal axes of known objects to compute objectness scores, enabling more accurate pseudo-label assignment for unknown objects. Since known and unknown object embeddings share inherent properties (\eg, objectness) absent in non-object proposals, we assume that unknown proposals projected onto the principal axes of known proposals exhibit similarity, whereas non-object proposals remain dissimilar. Based on this assumption, PAUL leverages these principal axes to identify object proposals and assign pseudo-labels to unknown objects.
Through the experiments shown in~\Cref{fig:motiv}, the proposed CollaPAUL can mitigate misclassification by knowledge confusion and detect unknown objects from the background, enhancing both known mAP and unknown recall performances (\ie, achieving the best H-Score).

\noindent
Our contributions can be summarized as follows:
\begin{itemize}
    \item We propose \textbf{SFUOD}, a new scenario where a source-trained detector adapts its knowledge to a target domain for known objects while simultaneously detecting unknown objects in the target domain. To this end, we introduce \textbf{CollaPAUL}, a novel framework combining collaborative tuning and principal axis-based unknown labeling.
    
    \item We propose \textit{collaborative tuning} to alleviate knowledge confusion by source-dependent knowledge. Through cross-domain attention, it integrates both source- and target-dependent knowledge, mitigating source reliance by collaboratively tuning target-dependent features.
    \item We propose a \textit{principal axis-based unknown labeling} method to assign pseudo-labels to unknown objects. It achieves this by projecting embeddings onto the principal axes of known objects to estimate objectness and computing confidence scores for unknown classes.
    \item We develop two SFUOD benchmarks, where the proposed CollaPAUL achieves state-of-the-art. We further validate its effectiveness through extensive experiments.
\end{itemize}

\section{Related Work}
\label{sec:related}

\subsection{Source-Free Object Detection}
Source-Free Object Detection (SFOD), introduced in SED~\cite{li2021free}, addresses domain adaptation with data privacy constraints, where the model adapts to the target domain without access to source data. Recent SFOD methods rely on the mean teacher framework for its robustness and stability, but noisy pseudo-labels from the teacher model can lead to the model collapse representing inaccurate student learning and improper teacher updates. To address this, PET~\cite{liu2023periodically} introduces a multi-teacher framework with static and dynamic teachers, stabilizing training by periodically exchanging the teacher and student models. SF-UT~\cite{hao2024simplifying} uses fixed pseudo-labels from the initial teacher model with AdaBN~\cite{li2018adaptive} to prevent improper teacher updates. DRU~\cite{khanh2024dynamic} dynamically retrains the student model and updates the teacher accordingly. While these methods focus on stabilizing teacher updates for source-free domain adaptation, we propose collaborative tuning, which improves student model learning by integrating source- and target-dependent representation knowledge for more effective adaptation.

\subsection{Open Recognition in Object Detection}
Open recognition has gained significant attention in object detection due to its importance in real-world applications. Unknown object detection~\cite{han2022expanding,liang2023unknown} trains a detector to classify annotated known objects while identifying unannotated novel objects as unknown. Open-world object detection (OWOD)~\cite{joseph2021towards,gupta2022ow,zohar2023prob,sun2024exploring} extends this by incorporating continual learning, requiring the detector to sequentially learn annotated known objects and detect unknowns. In domain adaptive object detection, SOMA~\cite{li2023novel} introduces Adaptive Open-set Object Detection (AOOD), which adapts a source-trained detector to a target domain, enabling it to classify known objects from the source while detecting unknowns. However, AOOD relies on source data for adaptation, raising privacy concerns. To address this, we propose Source-Free Unknown Object Detection (SFUOD), which adapts a source-trained model to the target domain without requiring source data. SFUOD enables the model to recognize known objects defined in the source domain while detecting novel objects as unknowns in the target domain.
\section{Method}
\label{sec:method}

\begin{figure*}[t]
    \centering
    \includegraphics[width=\linewidth]{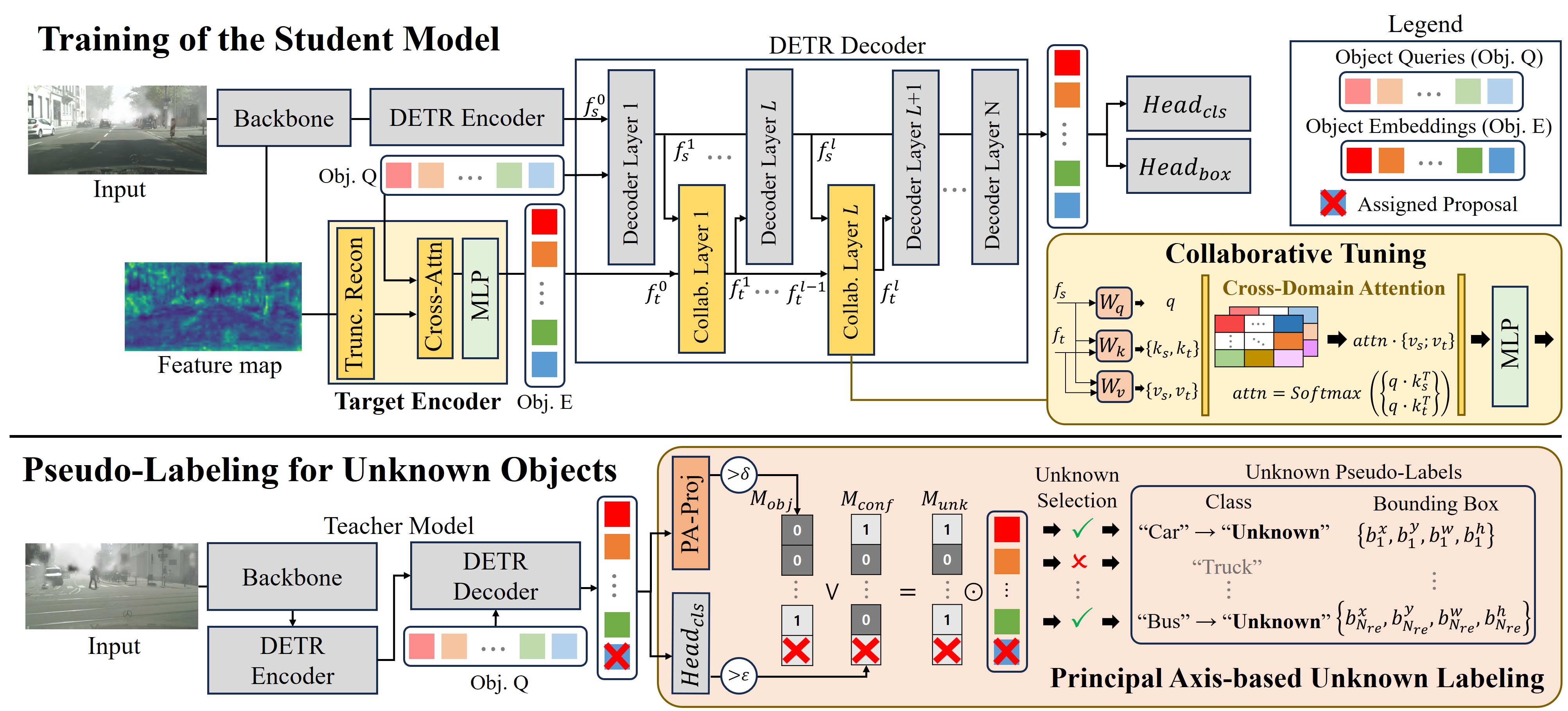}
    \vspace{-7mm}
    \caption{Overview of the proposed CollaPAUL framework. CollaPAUL consists of collaborative tuning and principal axis-based unknown labeling. The proposed collaborative tuning employs an auxiliary target encoder to extract target-dependent knowledge, integrated with source-dependent knowledge from the DETR encoder via cross-domain attention. A collaborative layer is inserted between the decoder layers to convey the integrated knowledge to the decoder. The proposed principal axis-based unknown labeling generates pseudo-labels for unknown objects by creating an objectness mask and confidence mask, selecting feature proposals representing unknown objects, and assigning the unknown label to them.\vspace{-5mm}}
    \label{fig:main}
\end{figure*}

We propose CollaPAUL, which incorporates collaborative tuning to mitigate knowledge confusion during knowledge transfer and principal axes-based unknown labeling to handle unknown objects. In~\cref{sec:prob_define}, we first define the source-free unknown object detection problem. \cref{sec:basemodel} describes the base model pipeline, followed by~\cref{sec:collabo}, where we introduce collaborative tuning that fuses source- and target-dependent knowledge through the cross-domain attention, improving adaptation to the target domain. Finally, in~\cref{sec:paul}, we present principal axes-based unknown labeling, assigning pseudo-labels to unknown objects by leveraging the principal axes of known objects to estimate the objectness. \Cref{fig:main} shows an overview of the CollaPAUL.

\subsection{Problem Definition}
\label{sec:prob_define}

In this section, we define the problem setting for the proposed SFUOD. Let 
$D_s = \{(x_s^i, y_s^i)\}_{i=1}^{|D_s|}$
denote the labeled source dataset and 
$D_t = \{(x_t^i)\}_{i=1}^{|D_t|}$
the unlabeled target dataset. Each source label 
$y_s^i = \{(b^j, c^j)\}_{j=1}^{J_i}$
contains the number of \( J_i \) annotations for known objects, where \( b^j \) represents the bounding box and \( c^j \in \mathcal{Y}_s \) denotes the object category for the $j$-th annotation. Unlike SFOD, which assumes a shared class space between source and target domains, SFUOD defines a set of known classes 
$\mathcal{Y}_s = \{1,2,\dots,K\}$
pre-defined in the source dataset and novel classes 
$\mathcal{Y}_t = \{K+2,K+3,\dots,K+K'\}$
that are undefined in the source dataset. In the source domain, all labeled objects belong to the base classes \( c^j \in \mathcal{Y}_s \), while the target domain may contain both base and novel classes. 
SFUOD aims to adapt the detector pre-trained on $D_s$ to the target domain using only $D_t$, enabling it to detect base classes while identifying novel classes as a single `unknown' category, represented as the class \( K+1 \) in the target domain.

\subsection{Pipeline of the Mean Teacher-based Approach}
\label{sec:basemodel}
We utilize the Mean Teacher (MT) framework~\cite{tarvainen2017mean}, which consists of a teacher model and a student model sharing the same architecture and initialization. The student model is trained using pseudo-labels generated by the teacher model, while the teacher model is updated via EMA from the student model. During training, the student and teacher models receive strongly and weakly augmented images, respectively. The updates for both models are as follows:
\begin{equation}
    \mathcal{L}_{det}=\mathcal{L}_{cls}(\hat{c}_{s},\hat{c}_{t}) + \mathcal{L}_{\mathrm{L}_1}(\hat{b}_{s},\hat{b}_{t}) + \mathcal{L}_{giou}(\hat{b}_{s},\hat{b}_{t}),
\end{equation}
\begin{equation}
\Bar{\theta}_t\leftarrow\alpha\theta_t + (1-\alpha)\theta_s,
\end{equation}
where $\mathcal{L}_{cls}$ denote the classification loss~\cite{lin2017focal}, $\mathcal{L}_{\mathrm{L}_1}$ and $\mathcal{L}_{giou}$ represent the regression losses~\cite{rezatofighi2019generalized}, and $\alpha$ denotes a hyperparameter for weight update.
The predicted class and bounding boxes from the student model \( \theta_s \) and teacher model \( \theta_t \) are denoted as \( \hat{c}_s, \hat{c}_t \) and \( \hat{b}_s, \hat{b}_t \), respectively. The teacher model \( \Bar{\theta}_t \) is updated using EMA and the student model \( \theta_s \) is trained via $\mathcal{L}_{det}$. The MT framework ensures consistent predictions during training, facilitating knowledge transfer to the target domain.

Most SFOD approaches leverage the MT framework to enhance adaptation while addressing improper model updates caused by noisy pseudo-labels. However, in the proposed SFUOD, the MT framework faces knowledge confusion, as both the student and teacher models, initialized with source-trained weights, struggle to learn representation knowledge for unknown objects in the target data. Unlike SFOD, SFUOD requires the model to not only adapt source knowledge for known objects but also learn effective representation knowledge for unknown objects in the target domain.

\subsection{Collaborative Tuning}
\label{sec:collabo}
The proposed SFUOD scenario requires the model to transfer the learned source knowledge to the target domain and simultaneously learn novel classes as unknown objects. Since unknown objects are not observed in the source domain, the adapted source knowledge hinders the model from learning representation knowledge for unknown objects, causing knowledge confusion between known and unknown objects. To this end, we propose collaborative tuning that aims to incorporate source-dependent knowledge extracted by the student model and target-dependent knowledge by an auxiliary target encoder to preserve transferable source knowledge while capturing representation knowledge for the target domain.

We first construct the target encoder, which consists of truncated reconstruction, a cross-attention layer, and an MLP layer. Given a $C$-dimensional feature map $f$ with spatial size $h\times w$ from the backbone (\eg, Resnet~\cite{he2016deep}), we calculate the magnitude of the feature activation $A\in \mathbb{R}^{h\times w}$ by averaging the feature map $f$ over the channels $C$ as in OW-DETR~\cite{gupta2022ow}. Then, we select top-$k$ activated features $f_a \in \mathbb{R}^{k \times C}$ from the~$A$.
We apply Singular Value Decomposition (SVD) like $f_a=U\Sigma V^T$, where $U \in \mathbb{R}^{k\times k}$, $\Sigma \in \mathbb{R}^{k\times C}$, and $V^T\in \mathbb{R}^{C\times C}$ denote the decomposed components, respectively, and reconstruct the feature map $f'_a=U'\Sigma' {V'}^T$, where $U'\in \mathbb{R}^{k \times r}, \Sigma'\in \mathbb{R}^{r \times r}$ and $V'^T\in \mathbb{R}^{r \times C}$, using the top-$r$ decomposed components to explore latent knowledge of the target domain. Since the feature $f'_a$ is reconstructed using only the principal components that encapsulate sufficient knowledge, it effectively reveals latent representations. The reconstructed feature map $f'_a$ then undergoes cross-attention with object queries $q_{obj}$ and MLP layers to extract target domain representations $f_t$ and align the dimension of the feature vector for the decoder layers.
Since the target encoder operates independently from the source-trained encoder in the student model, it learns target-dependent, source-independent knowledge, capturing effective representations for the target domain.

To integrate target-dependent knowledge with source-dependent knowledge, we feed target-dependent features $f_t$ into the decoder layers of the student model, defined as $\theta_s^{dec}=\{\theta_s^{dec_l}\}_{l=1}^{|N_{dec}|}$, where $|N_{dec}|$ denotes the number of decoder layers. To enable collaborative learning between the source-dependent knowledge $f_s$ from the student model and the target-dependent knowledge $f_t$ from the target encoder, we introduce collaborative layers, consisting of a cross-domain attention layer and an MLP layer. The cross-domain attention computes attention scores by aligning source and target features using identical object queries, facilitating effective knowledge integration. Given $q=f_s\cdot W_q^T$, $k=[f_s;f_t]\cdot W_k^T$, and $v=[f_s;f_t]\cdot W_v^T$, where $q \in \mathbb{R}^{N_q \times D}$, $k \in \mathbb{R}^{2N_q \times D}$, and $v \in \mathbb{R}^{2N_q \times D}$ denote the query, key, and value for the attention layer, respectively, $N_q$ denotes the number of object queries, and $D$ denotes the dimension of the feature vectors, the cross-domain attention is calculated as follows:
\begin{align}
    attn=& \sigma\left(\begin{Bmatrix} q\cdot k_s^T \\ q\cdot k_t^T \end{Bmatrix}\right) \in \mathbb{R}^{N_q \times 2 \times N_q}, \label{eq:3}\\
    f_{cd}=& \,attn \cdot \{v_s; v_t\} \in \mathbb{R}^{N_q\times D}, \label{eq:4} 
\end{align}
where $\sigma(\cdot)$ and $f_{cd}$ denote the softmax function and the output of the cross-domain attention layer. In \cref{eq:3},
We compute the softmax score of each query on the corresponding source- and target-dependent features and concatenate $attn \in \mathbb{R}^{N_q \times 2N_q}$ to compute the attention value, as in~\Cref{eq:4}. Then, we feed \( f_{cd} \) into the MLP layers to refine the integrated knowledge.


As shown in~\Cref{fig:main}, to convey integrated knowledge by the collaborative (collab.) layer to the target decoder, we insert a collaborative layer between the decoder layers.
Specifically, the collaborative layer is inserted into the first \( L \) decoder layers, excluding the initial layer, as its output feature is used for source-dependent representations. Here, \( L \) is a hyperparameter that determines the number of inserted layers. We set \( L = 3 \) based on empirical analysis.
Using the first output from the decoder layer $f_s^1=\theta_s^{dec_1}(f_{s}^{0})$ and the output of target encoder $f_t^0$, where $f_s^0$ denotes the output of the student encoder, propagations of the $(l+1)$-th decoder and the $l$-th collaborative layer for collaborative tuning are calculated as follows:
\begin{align}
    f_t^l=&\;\theta^{clb_l}(f_s^{l},f_t^{l-1}), \\
    f_s^{l+1}=&\;\theta_s^{dec_{l+1}}([f_{s}^{l};f_t^{l}]),
\end{align}
where $\theta^{clb_l}$ and $\theta^{dec_{l+1}}$ denote weights for the collaborative layer and decoder layer, respectively.
During the decoder layer, which consists of self-attention and cross-attention, we apply prefix tuning~\cite{li2021prefix} to feed \( f_t^l \) into the self-attention layer, leveraging its effectiveness in knowledge adaptation. 
Through the repeated propagation of the decoder and collaborative layers, the collaborative tuning trains the collaborative layer to integrate source- and target-dependent knowledge, while the decoder extracts enhanced representations using features from both the previous layer and the collaborative layer.
The collaborative tuning enhances representation learning for the target domain and facilitates knowledge transfer from source to target domain.

\subsection{Principal Axis-based Unknown Labeling}
\label{sec:paul}
The SFUOD scenario requires the model to not only classify known objects but also identify novel objects as unknown, which are not defined in the source domain. However, due to the absence of knowledge of unknown objects in the source-trained model, the teacher model in the mean teacher framework struggles to assign reliable pseudo-labels to unknown objects. As a result, the detector often fails to detect unknown objects in SFUOD. To address this, we propose Principal Axis-based Unknown Labeling (PAUL), which estimates the objectness of proposal features and leverages an unknown confidence score from the prediction to effectively select proposal features for assigning unknown pseudo-labels.

The input data $x_t$ is fed into the teacher model to extract \( N_q \) proposal features and generate predictions, including class labels and bounding boxes. First, pseudo-labels are assigned to \( N_k \) known proposals using a confidence-based approach leveraging the predicted confidence scores with the threshold, \eg, we use 0.3 for all experiments. Then, to assign unknown pseudo-labels among the $N_q-N_k$ remaining proposals, the proposed PAUL is applied to filter unknown proposals and assign pseudo-labels accordingly.

\begin{table*}[t]
\resizebox{\linewidth}{!}{%
\setlength{\tabcolsep}{12pt} 
\renewcommand{\arraystretch}{1.} 
\begin{tabular}{cc|ccc|cgy}
\specialrule{1pt}{0pt}{0pt}
\multirow{2}{*}{Setting} & \multirow{2}{*}{Methods} & \multicolumn{6}{c}{Cityscape $\rightarrow$ Foggy Cityscape} \\ \cline{3-8}
 &  & Car & Truck & Bus & Known mAP & U-Recall & H-Score \\ \hline
 & Source only & 43.20 & 12.05 & 24.43 & 26.56 & 0.00 & 0.00 \\ \hline
\multirow{4}{*}{SFOD} & Mean Teacher~\cite{tarvainen2017mean} & \underline{50.20} & 0.00 & 0.54 & 16.91 & \underline{6.02} & \underline{8.88} \\
 & PET~\cite{liu2023periodically} & 36.38 & 1.45 & 0.00 & 12.61 & 2.76 & 4.53 \\
 & DRU~\cite{khanh2024dynamic} & 41.14 & 9.65 & 18.12 & 22.97 & 3.60 & 6.22 \\
 & SF-UT~\cite{li2021free} & 39.82 & \underline{13.83} & \underline{24.90} & \underline{26.18} & 0.00 & 0.00 \\ \hdashline
SFUOD & \textbf{CollaPAUL (Ours)} & \textbf{52.10} & \textbf{16.49} & \textbf{28.37} & \textbf{32.32} & \textbf{10.59} & \textbf{15.95} \\
\specialrule{1pt}{0pt}{0pt}
\end{tabular}%
}
\vspace{-3mm}
\caption{Experimental results on the weather adaptation benchmark (Cityscapes $\rightarrow$ Foggy Cityscapes). ``Source only'' refers to the model trained solely on the source domain. The best and second-best performances are highlighted in bold and underlined, respectively. \vspace{-2mm}}
\label{tab:main_foggy}
\end{table*}

\begin{table*}
\resizebox{\linewidth}{!}{%
\setlength{\tabcolsep}{12pt} 
\renewcommand{\arraystretch}{1.} 
\begin{tabular}{cc|ccc|cgy}
\specialrule{1pt}{0pt}{0pt}
\multirow{2}{*}{Setting} & \multirow{2}{*}{Methods} & \multicolumn{6}{c}{Cityscape $\rightarrow$ BDD100K} \\ \cline{3-8}
 &  & Car & Truck & Bus & Known mAP & U-Recall & H-Score \\ \hline
 & Source only & 51.38 & 13.73 & 8.13 & 24.41 & 0.00 & 0.00 \\ \hline
\multirow{4}{*}{SFOD} & Mean Teacher~\cite{tarvainen2017mean} & \textbf{59.68} & \underline{13.04} & 6.55 & 26.42 & 6.08 & \underline{9.89} \\
 & PET~\cite{liu2023periodically} & 48.56 & 6.54 & 2.26 & 19.12 & 1.98 & 3.59 \\
 & DRU~\cite{khanh2024dynamic} & 44.91 & 4.14 & 2.68 & 17.24 & \underline{6.86} & 9.82 \\
 & SF-UT~\cite{li2021free} & 52.51 & 12.31 & \textbf{15.21} & \underline{26.68} & 1.17 & 2.24 \\ \hdashline
SFUOD & \textbf{CollaPAUL (Ours)} & \underline{57.95} & \textbf{17.27} & \underline{9.40} & \textbf{28.21} & \textbf{8.57} & \textbf{13.15} \\
\specialrule{1pt}{0pt}{0pt}
\end{tabular}%
}
\vspace{-3mm}
\caption{Experimental results on the cross-scene adaptation benchmark (Cityscapes $\rightarrow$ BDD100K).``Source only'' refers to the model trained solely on the source domain. The best and second-best performances are highlighted in bold and underlined, respectively. \vspace{-4mm}}
\label{tab:main_bdd}
\end{table*}

Since the principal axes, computed from assigned known proposal features, represent the direction that preserves representation knowledge, we assume that unknown proposals projected onto the principal axes of known proposals exhibit similarity to projected known proposals due to sharing objectness, while non-object proposals remain dissimilar. We project both known and remaining proposals onto the principal axes to compute the objectness scores via cosine similarity. 
Given the $\Bar{f}_{kn}=f_{kn}\cdot P^T$, $\Bar{f}_{re}=f_{re}\cdot P^T$, where $k_{kn}$, $f_{re}$, and $P$ denote known proposals, remaining proposals, and principals axes, respectively, the objectness score is formulated as follows:
\vspace{-3mm}
\begin{align}
    s_{kn}^{i}=&\frac{1}{N_k-1}\sum_{j=1, j\neq i}^{N_k}d(\Bar{f}_{kn}^i,\Bar{f}_{kn}^j), \\
    s_{re}^{i}=&\frac{1}{N_{re}}\sum_{j=1}^{N_{re}}d(\Bar{f}_{re}^i,\Bar{f}_{kn}^j),
\end{align}
where \( \Bar{f}_{kn}^i \) and \( \Bar{f}_{re}^i \) represent the \( i \)-th known and remaining proposals projected onto the principal axes, respectively, \( N_{re} = N_q - N_k \) denotes the number of remaining proposals, and \( d(\cdot,\cdot) \) represents cosine similarity. The threshold \( \delta \) is set as the average of the known objectness scores $S_{kn}=\{s_{kn}^1, s_{kn}^2,\dots, s_{kn}^{N_k}\}$, and the objectness mask $M_{obj}$ is constructed by comparing the remaining scores with the threshold \( \delta \). Additionally, a confidence mask $M_{conf}$ is generated based on the confidence scores of the teacher model, using a confidence threshold \( \epsilon \).
Finally, an unknown mask $M_{unk}$ is generated by merging the objectness and confidence masks to select reliable unknown proposals. This mask ensures that selected proposals exhibit shareable objectness while representing unknown classes.
The process of selecting unknown proposals is as follows:
\begin{equation}
    M_{obj}= \{m_{obj}^i|m_{obj}^i=\mathbb{I}(s_{un}^i\geq \delta)\}_{i=1}^{N_{re}}, 
\end{equation}
\begin{equation}
    M_{conf}= \{m_{obj}^j|m_{obj}^j=\mathbb{I}(c_{un}^j\geq \epsilon)\}_{j=1}^{N_{re}},
\end{equation}
\begin{equation}
    M_{unk}= M_{obj} \vee M_{conf},
\end{equation}
where $\mathbb{I}(\cdot)$ denotes the indicator function, $c_{un}^j$ denotes the unknown confidence score predicted for the $j$-th remaining proposal, and $\delta$ and $\epsilon$ denote the objectness and confidence thresholds, respectively. 
We apply an element-wise product between the remaining proposals and the unknown mask to identify activated proposals as unknown objects. Pseudo-labels, including class labels and bounding boxes, are then assigned based on the predictions of the teacher model for the selected proposals.

To sum up, we propose the CollaPAUL framework, which integrates collaborative tuning and principal axis-based unknown labeling. Collaborative tuning mitigates knowledge confusion caused by source-dependent knowledge, which hinders effective knowledge transfer. Integrating source- and target-dependent knowledge, 
collaborative tuning enables the model to enhance knowledge adaptation and improves learning of representations of the target domain.
Additionally, we introduce an effective unknown pseudo-labeling to address detection failures for unknown objects. This approach selects unknown proposals based on objectness and confidence scores, enabling the model to detect unknown objects more effectively during training.
\section{Experiments}
\label{sec:experiments}

\subsection{Experimental Settings}

\paragraph{Benchmarks.} 
To benchmark the proposed SFUOD, we utilized the existing SFOD benchmarks: weather adaptation and cross-scene adaptation.
In weather adaptation, Cityscapes~\cite{cordts2016cityscapes} served as the source domain, while Foggy Cityscapes~\cite{sakaridis2018semantic} (fog density 0.02) was the target domain. Cityscapes consists of 3,475 annotated urban images, with 2,975 for training and 500 for evaluation, while Foggy Cityscapes is generated via fog synthesis. In cross-scene adaptation, the BDD100K~\cite{yu2020bdd100k} daytime subset was used as the target domain, containing 36,728 training and 5,258 validation images with annotations, while Cityscapes remained the source domain. For both benchmarks, vehicle classes (\eg, ``Car'', ``Truck'', ``Bus'']) were predefined as known classes, while other categories (\eg, ``Person'', ``Rider'', ``Motorcycle'', ``Train'', ``Bicycle'') were treated as unknown objects.
\vspace{-7mm}
\paragraph{Metrics.}
We validated CollaPAUL by evaluating both known and unknown classes. Known class performance was measured using mean average precision (known mAP), which quantifies mAP for predefined known classes in the target domain. Unknown class detection was assessed using recall (U-Recall). To evaluate overall performance across both categories, we computed the harmonic mean (H-score), which balances known mAP and U-Recall.
\vspace{-7mm}
\paragraph{Implementation Details.}
We used DRU~\cite{khanh2024dynamic} as the base model, as it effectively manages student learning and teacher updates within the Mean Teacher framework. For the detector, we adopted Deformable-DETR~\cite{zhudeformable} with a ResNet-50 backbone pre-trained on ImageNet. CollaPAUL was trained using the AdamW optimizer, with a known object pseudo-labeling threshold and an unknown threshold $\delta$ for the proposed PAUL set to 0.3. We set $\alpha$ to 0.99 for the EMA updates. Training was conducted on 4 NVIDIA RTX 3090 GPUs, with a batch size of 2 per GPU.

\begin{table}
\resizebox{\columnwidth}{!}{%
\setlength{\tabcolsep}{6pt} 
\renewcommand{\arraystretch}{1.} 
\begin{tabular}{cc|cgy}
\specialrule{1pt}{0pt}{0pt}
\multicolumn{2}{c}{Ablation} & \multicolumn{3}{|c}{Cityscapes $\rightarrow$ Foggy Cityscapes} \\ \hline
Collab & PAUL & Known mAP & U-Recall & H-Score \\ \hline
 &  & 22.97 & 3.60 & 6.22 \\ \hdashline
\checkmark &  & \underline{30.63} & 3.56 & 6.38 \\
 & \checkmark & 25.40 & \underline{6.46} & \underline{10.30} \\ \hdashline
\checkmark & \checkmark & \textbf{32.32} & \textbf{10.59} & \textbf{15.95}\\
\specialrule{1pt}{0pt}{0pt}
\end{tabular}%
}
\vspace{-2mm}
\caption{Ablation study of CollaPAUL on weather adaptation. Collab and PAUL denote the proposed collaborative tuning and principal axes-based unknown labeling.}
\label{tab:ablation}
\vfill
\resizebox{\columnwidth}{!}{%
\setlength{\tabcolsep}{4pt} 
\renewcommand{\arraystretch}{1.} 
\begin{tabular}{c|cgy}
\specialrule{1pt}{0pt}{0pt}
\multirow{2}{*}{\# of Collab Layers} & \multicolumn{3}{c}{Cityscapes $\rightarrow$ Foggy Cityscapes} \\ \cline{2-4}
 & Known mAP & U-Recall & H-Score \\ \hline
$L=$ 1 & 22.63 & 7.23 & 10.96 \\
$L=$ 2 & 26.98 & 7.25 & 11.43 \\ \hdashline
$\boldsymbol{L=}$ \textbf{3} & \underline{32.32} & \textbf{10.59} & \textbf{15.95} \\ \hdashline
$L=$ 4 & \textbf{32.90} & 7.44 & 12.14 \\
$L=$ 5 & 30.33 & \underline{8.29} & \underline{13.02} \\
\specialrule{1pt}{0pt}{0pt}
\end{tabular}%
}
\vspace{-2mm}
\caption{Ablation study for the number of inserted collaborative layers on weather adaptation.\vspace{-5mm}}
\label{tab:ablation_layer}
\end{table}

\subsection{Experimental Results}
In this section, we evaluated CollaPAUL on the Weather Adaptation (Cityscapes $\rightarrow$ Foggy Cityscapes) and Cross-Scene Adaptation (Cityscapes $\rightarrow$ BDD100K) benchmarks and compared its performance with recent source-free object detection methods using the Mean Teacher framework.
\vspace{-9mm}
\paragraph{Weather Adaptation.}
Weather adaptation captures real-world variations in weather conditions. To assess detector robustness, we conducted experiments on Cityscapes $\rightarrow$ Foggy Cityscapes. As shown in~\Cref{tab:main_foggy}, CollaPAUL outperformed all baselines across all metrics. Notably, the mean teacher framework outperformed existing SFOD methods based on the mean teacher. Since SFOD methods focus on stabilizing teacher updates, they update the teacher model slowly, limiting its ability to learn unknown objects and restricting the student model from effectively utilizing pseudo-labels. 
In contrast, CollaPAUL achieved performance gains of up to 6.14\% in known mAP, 4.57\% in U-Recall, and 7.07\% in H-Score, demonstrating its robustness to weather changes.
\vspace{-5mm}
\paragraph{Cross-Scene Adaptation.}
Scene configurations dynamically change in different locations in the real world, especially in automated driving contexts. Hence, cross-scene adaptation is crucial in domain adaptation. As shown in~\Cref{tab:main_bdd}, the proposed CollaPAUL surpassed the baselines across all the metrics. It gained performance enhancements of 1.53\% in known mAP, 1.71\% in U-Recall, and 3.26\% in H-Score. This demonstrates the effectiveness of the proposed CollaPAUL in cross-scene adaptation.

\subsection{Ablation Study}
We conducted an ablation study on the weather adaptation benchmark to validate the performance improvements of the proposed method. Additionally, we analyzed the impact of collaborative tuning by varying the number of inserted collaborative layers and further examined the proposed principal axes-based unknown labeling.
\vspace{-5mm}
\paragraph{Ablation for the CollaPAUL.}
Collab and PAUL represent the proposed collaborative tuning and principal axes-based unknown labeling, with DRU as the baseline. As shown in~\Cref{tab:ablation}, collaborative learning improved known mAP by 7.66\%, while PAUL enhanced U-Recall by 2.86\%. Their combination, CollaPAUL, achieved strong performance, showing that proposed components enhance knowledge transfer for known objects and facilitate unknown pseudo-labeling, leading to synergistic improvements.
\vspace{-5mm}
\begin{table}
\resizebox{\columnwidth}{!}{%
\setlength{\tabcolsep}{6pt} 
\renewcommand{\arraystretch}{1.} 
\begin{tabular}{cc|cgy}
\specialrule{1pt}{0pt}{0pt}
\multicolumn{2}{c}{PAUL} & \multicolumn{3}{|c}{Cityscapes $\rightarrow$ Foggy Cityscapes} \\ \hline
$M_{conf}$ & $M_{obj}$ & Known mAP & U-Recall & H-Score \\ \hline
\checkmark &  & \underline{30.63} & 3.56 & 6.38 \\
 & \checkmark & 29.88 & \underline{9.65} & \underline{14.59} \\ \hdashline
\checkmark & \checkmark & \textbf{32.32} & \textbf{10.59} & \textbf{15.95} \\
\specialrule{1pt}{0pt}{0pt}
\end{tabular}%
}
\vspace{-2mm}
\caption{Ablation study of the proposed PAUL, which consists of confidence and objectness mask on weather adaptation.}
\label{tab:ablation_PAUL}
\vfill
\resizebox{\columnwidth}{!}{%
\setlength{\tabcolsep}{4pt} 
\renewcommand{\arraystretch}{1.} 
\begin{tabular}{c|cgy}
\specialrule{1pt}{0pt}{0pt}
\multirow{2}{*}{Collab Tuning} & \multicolumn{3}{c}{Cityscapes $\rightarrow$ Foggy Cityscapes} \\ \cline{2-4}
 & Known mAP & U-Recall & H-Score \\ \hline
Baseline (only PAUL) & 25.40 & 6.46 & 10.30 \\ \hdashline
Prefix-tuning & \underline{28.43} & \underline{8.07} & \underline{12.57} \\
\textbf{Cross-domain (Ours)} & \textbf{32.32} & \textbf{10.59} & \textbf{15.95} \\
\specialrule{1pt}{0pt}{0pt}
\end{tabular}%
}
\vspace{-2mm}
\caption{Analysis of the proposed cross-domain attention. Baseline denotes employing on PAUL.}
\label{tab:analysis_collab}
\vfill
\resizebox{\columnwidth}{!}{%
\setlength{\tabcolsep}{4pt} 
\renewcommand{\arraystretch}{1.} 
\begin{tabular}{c|cgy}
\specialrule{1pt}{0pt}{0pt}
\multirow{2}{*}{Unknown Pseudo Label} & \multicolumn{3}{c}{Cityscapes $\rightarrow$ Foggy Cityscapes} \\ \cline{2-4}
 & Known mAP & U-Recall & H-Score \\ \hline
Confidence-based & 30.63 & 3.56 & 6.38 \\ \hdashline
Attention-driven & \underline{32.18} & \underline{4.00} & \underline{7.12} \\
\textbf{PAUL (Ours)} & \textbf{32.32} & \textbf{10.59} & \textbf{15.95} \\
\specialrule{1pt}{0pt}{0pt}
\end{tabular}%
}
\vspace{-2mm}
\caption{Analysis of the principal axes-based unknown labeling. We used confidence-based unknown labeling as a baseline.\vspace{-7mm}}
\label{tab:analysis_paul}
\end{table}

\begin{figure*}[t]
    \centering
    \includegraphics[width=\linewidth]{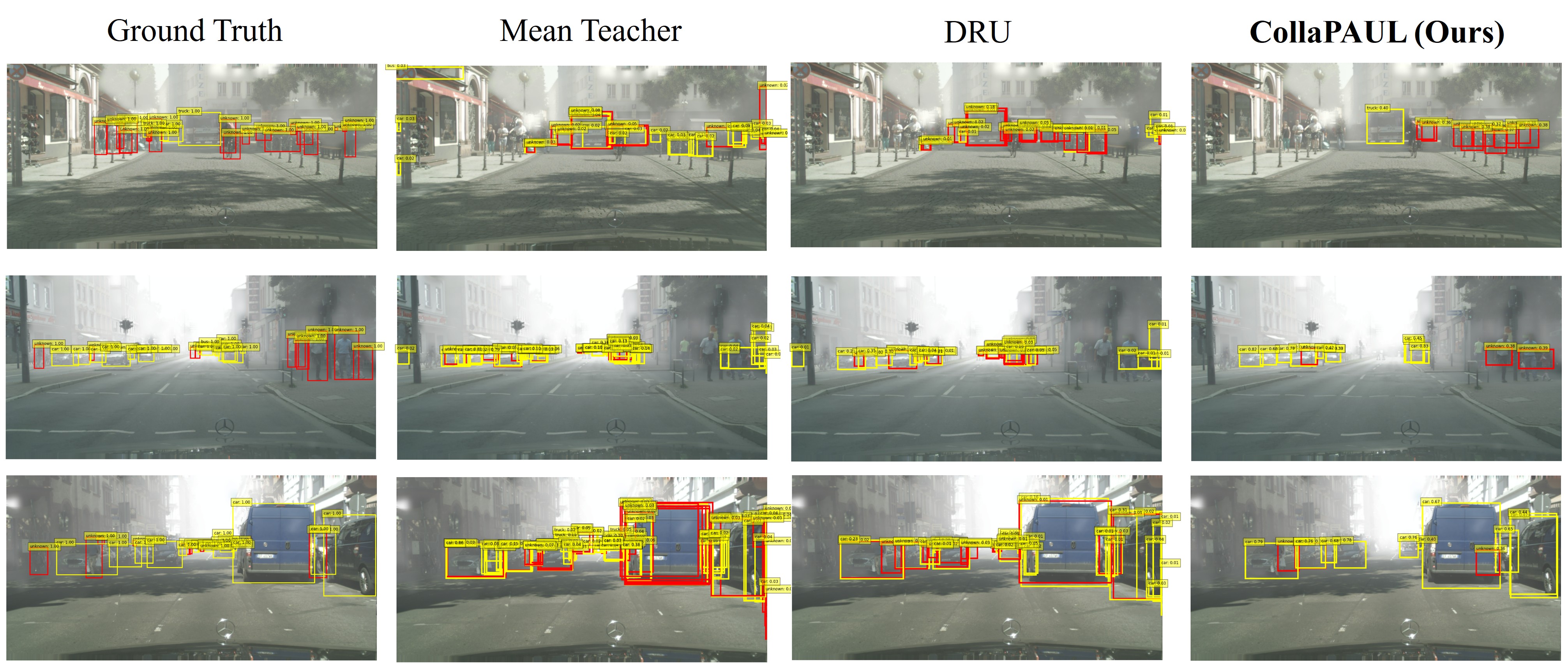}
    \vspace{-6mm}
    \caption{Qualitative results on weather adaptation. The visualizations compare the performance of Mean Teacher, DRU, and CollaPAUL, with known objects highlighted in yellow boxes and unknown objects in red boxes.\vspace{-6mm}}
    \label{fig:qual_viz}
\end{figure*}

\paragraph{Ablation for the Collaborative Tuning.}
We conducted an ablation study to determine the optimal number of collaborative layers for the proposed collaborative tuning. To ensure stable propagation through the decoders, we inserted these layers from the initial stages. As shown in~\Cref{tab:ablation_layer}, inserting three layers showed the best U-Recall and H-Score. While performance improved with up to three layers, adding more resulted in a declined performance.
\vspace{-5mm}
\paragraph{Ablation for the PAUL.}
We further examined the effectiveness of the proposed principal axes-based unknown labeling, which includes confidence and objectness masks. Confidence-based labeling, widely used in open-set learning, served as the baseline. As shown in~\Cref{tab:ablation_PAUL}, employing the objectness mask significantly improved unknown object detection by 6.09\%, while incorporating both masks achieved the best overall performance. This demonstrates that principal axes-based objectness estimation enhances unknown object detection.

\subsection{Analysis}
\paragraph{Analysis for the Cross-domain Attention.}
To validate the effectiveness of the proposed cross-domain attention, designed to integrate source- and target-dependent knowledge in the collaborative layer, we compared its performance against prefix-tuning, a widely used parameter-efficient tuning method. As shown in~\Cref{tab:analysis_collab}, with PAUL as the baseline, prefix-tuning, which integrates attention knowledge from all feature vectors, improved the performance of the source-trained model. However, our cross-domain attention, which focuses on integrating source- and target-dependent knowledge extracted from identical object queries, achieved superior results across all metrics, with increases of 3.89\% in known mAP, 2.52\% in U-Recall, and 3.38\% in H-Score. This validates its effectiveness in integrating domain-dependent representation knowledge.
\vspace{-5mm}
\paragraph{Comparison of the Unknown Labeling.}
To validate the effectiveness of the proposed principal axes-based unknown labeling (PAUL), we compared its performance with other unknown labeling methods. Specifically, we evaluated PAUL against confidence-based approaches (baseline) and attention-driven unknown labeling from OW-DETR~\cite{gupta2022ow}. As shown in~\Cref{tab:analysis_paul}, PAUL outperformed all methods, achieving significant improvements of 6.59\% in U-Recall and 8.83\% in H-Score. While attention-driven labeling from OW-DETR showed slight gains, PAUL demonstrated superior performance. This shows that PAUL effectively enhances unknown object detection by enabling the teacher model to generate high-quality unknown pseudo-labels.
\vspace{-5mm}
\paragraph{Qualitative Results on Weather Adaptation.}
To demonstrate the real-world applicability of our method, we conducted qualitative experiments comparing CollaPAUL with the mean teacher framework and the DRU method. We visualized results from the weather adaptation benchmark, where yellow boxes indicate known objects (\eg, ``Car'', ``Truck'', ``Bus'') and red boxes represent novel objects (\eg, ``Person'', ``Rider'', ``Motorcycle'', ``Train'', ``Bicycle'') predicted as unknowns. As shown in~\Cref{fig:qual_viz}, the mean teacher and DRU methods struggled with knowledge confusion, misclassifying known objects as unknown and vice versa, and failed to distinguish unknown objects from the background. In contrast, CollaPAUL effectively mitigated knowledge confusion, enabling the detector to correctly classify known and unknown objects while accurately identifying unknowns against the background. These results show the superior performance of CollaPAUL in the proposed source-free unknown object detection.

\vspace{-2mm}
\section{Conclusion}
\label{sec:conclusion}
In this paper, we propose a novel \textbf{S}ource-\textbf{F}ree \textbf{U}nknown \textbf{O}bject \textbf{D}etection (\textbf{SFUOD}) scenario, where the model recognizes known objects while detecting novel objects as unknowns. The proposed SFUOD presents a challenging yet realistic setting due to knowledge confusion and limited knowledge of unknowns.
To address this, we propose \textbf{Colla}borative tuning and \textbf{P}rincipal \textbf{A}xis-based \textbf{U}nknown \textbf{L}abeling (\textbf{CollaPAUL}).
The proposed CollaPAUL mitigates knowledge confusion through effective knowledge transfer via collaborative tuning and generates unknown pseudo-labels via principal axis-based unknown labeling, estimating objectness by utilizing principal axes of known objects.
While there is room for improvement in SFUOD, CollaPAUL outperforms prior SFOD methods.
We expect our research to offer novel insights into domain adaptive object detection with open-set recognition for more realistic applications.

\newpage
\section*{Acknowledgments}
\label{sec:ack}
This research was conducted with the support of the HANCOM InSpace Co., Ltd. (Hancom-Kyung Hee Artificial Intelligence Research Institute), and supported by Institute of Information \& communications Technology Planning \& Evaluation (IITP) grant funded by the Korea government (MSIT) (RS-2019-II190079, Artificial Intelligence Graduate School Program (Korea University), and RS-2024-00457882, AI Research Hub Project).

{
    \small
    \bibliographystyle{ieeenat_fullname}
    \bibliography{main}

\begin{thebibliography}{33}
\providecommand{\natexlab}[1]{#1}
\providecommand{\url}[1]{\texttt{#1}}
\expandafter\ifx\csname urlstyle\endcsname\relax
  \providecommand{\doi}[1]{doi: #1}\else
  \providecommand{\doi}{doi: \begingroup \urlstyle{rm}\Url}\fi

\bibitem[Chen et~al.(2018)Chen, Li, Sakaridis, Dai, and Van~Gool]{chen2018domain}
Yuhua Chen, Wen Li, Christos Sakaridis, Dengxin Dai, and Luc Van~Gool.
\newblock Domain adaptive faster r-cnn for object detection in the wild.
\newblock In \emph{Proceedings of the IEEE conference on computer vision and pattern recognition}, pages 3339--3348, 2018.

\bibitem[Choe et~al.(2024)Choe, Shin, Park, Choi, and Park]{choe2024open}
Seun-An Choe, Ah-Hyung Shin, Keon-Hee Park, Jinwoo Choi, and Gyeong-Moon Park.
\newblock Open-set domain adaptation for semantic segmentation.
\newblock In \emph{Proceedings of the IEEE/CVF Conference on Computer Vision and Pattern Recognition}, pages 23943--23953, 2024.

\bibitem[Chu et~al.(2023)Chu, Li, Chen, Li, and Li]{chu2023adversarial}
Qiaosong Chu, Shuyan Li, Guangyi Chen, Kai Li, and Xiu Li.
\newblock Adversarial alignment for source free object detection.
\newblock In \emph{Proceedings of the AAAI conference on artificial intelligence}, pages 452--460, 2023.

\bibitem[Cordts et~al.(2016)Cordts, Omran, Ramos, Rehfeld, Enzweiler, Benenson, Franke, Roth, and Schiele]{cordts2016cityscapes}
Marius Cordts, Mohamed Omran, Sebastian Ramos, Timo Rehfeld, Markus Enzweiler, Rodrigo Benenson, Uwe Franke, Stefan Roth, and Bernt Schiele.
\newblock The cityscapes dataset for semantic urban scene understanding.
\newblock In \emph{Proceedings of the IEEE conference on computer vision and pattern recognition}, pages 3213--3223, 2016.

\bibitem[Deng et~al.(2021)Deng, Li, Chen, and Duan]{deng2021unbiased}
Jinhong Deng, Wen Li, Yuhua Chen, and Lixin Duan.
\newblock Unbiased mean teacher for cross-domain object detection.
\newblock In \emph{Proceedings of the IEEE/CVF conference on computer vision and pattern recognition}, pages 4091--4101, 2021.

\bibitem[Deng et~al.(2023)Deng, Zhang, Li, Duan, and Xu]{deng2023cross}
Jinhong Deng, Xiaoyue Zhang, Wen Li, Lixin Duan, and Dong Xu.
\newblock Cross-domain detection transformer based on spatial-aware and semantic-aware token alignment.
\newblock \emph{IEEE Transactions on Multimedia}, 26:\penalty0 5234--5245, 2023.

\bibitem[Fu et~al.(2020)Fu, Cao, Long, and Wang]{fu2020learning}
Bo Fu, Zhangjie Cao, Mingsheng Long, and Jianmin Wang.
\newblock Learning to detect open classes for universal domain adaptation.
\newblock In \emph{Computer vision--ECCV 2020: 16th European conference, glasgow, UK, August 23--28, 2020, proceedings, part XV 16}, pages 567--583. Springer, 2020.

\bibitem[Gupta et~al.(2022)Gupta, Narayan, Joseph, Khan, Khan, and Shah]{gupta2022ow}
Akshita Gupta, Sanath Narayan, KJ Joseph, Salman Khan, Fahad~Shahbaz Khan, and Mubarak Shah.
\newblock Ow-detr: Open-world detection transformer.
\newblock In \emph{Proceedings of the IEEE/CVF conference on computer vision and pattern recognition}, pages 9235--9244, 2022.

\bibitem[Han et~al.(2022)Han, Ren, Ding, Pan, Yan, and Xia]{han2022expanding}
Jiaming Han, Yuqiang Ren, Jian Ding, Xingjia Pan, Ke Yan, and Gui-Song Xia.
\newblock Expanding low-density latent regions for open-set object detection.
\newblock In \emph{Proceedings of the IEEE/CVF Conference on Computer Vision and Pattern Recognition}, pages 9591--9600, 2022.

\bibitem[Hao et~al.(2024)Hao, Forest, and Fink]{hao2024simplifying}
Yan Hao, Florent Forest, and Olga Fink.
\newblock Simplifying source-free domain adaptation for object detection: Effective self-training strategies and performance insights.
\newblock In \emph{European Conference on Computer Vision}, pages 196--213. Springer, 2024.

\bibitem[He et~al.(2016)He, Zhang, Ren, and Sun]{he2016deep}
Kaiming He, Xiangyu Zhang, Shaoqing Ren, and Jian Sun.
\newblock Deep residual learning for image recognition.
\newblock In \emph{Proceedings of the IEEE conference on computer vision and pattern recognition}, pages 770--778, 2016.

\bibitem[Hsu et~al.(2020)Hsu, Tsai, Lin, and Yang]{hsu2020every}
Cheng-Chun Hsu, Yi-Hsuan Tsai, Yen-Yu Lin, and Ming-Hsuan Yang.
\newblock Every pixel matters: Center-aware feature alignment for domain adaptive object detector.
\newblock In \emph{Computer Vision--ECCV 2020: 16th European Conference, Glasgow, UK, August 23--28, 2020, Proceedings, Part IX 16}, pages 733--748. Springer, 2020.

\bibitem[Joseph et~al.(2021)Joseph, Khan, Khan, and Balasubramanian]{joseph2021towards}
KJ Joseph, Salman Khan, Fahad~Shahbaz Khan, and Vineeth~N Balasubramanian.
\newblock Towards open world object detection.
\newblock In \emph{Proceedings of the IEEE/CVF conference on computer vision and pattern recognition}, pages 5830--5840, 2021.

\bibitem[Khanh et~al.(2024)Khanh, Nguyen, Pham, Tran, and Jeon]{khanh2024dynamic}
Trinh Le~Ba Khanh, Huy-Hung Nguyen, Long~Hoang Pham, Duong Nguyen-Ngoc Tran, and Jae~Wook Jeon.
\newblock Dynamic retraining-updating mean teacher for source-free object detection.
\newblock In \emph{European Conference on Computer Vision}, pages 328--344. Springer, 2024.

\bibitem[Li et~al.(2022{\natexlab{a}})Li, Ye, Zhu, Zhou, and Xiong]{li2022source}
Shuaifeng Li, Mao Ye, Xiatian Zhu, Lihua Zhou, and Lin Xiong.
\newblock Source-free object detection by learning to overlook domain style.
\newblock In \emph{Proceedings of the IEEE/CVF conference on computer vision and pattern recognition}, pages 8014--8023, 2022{\natexlab{a}}.

\bibitem[Li et~al.(2022{\natexlab{b}})Li, Liu, and Yuan]{li2022sigma}
Wuyang Li, Xinyu Liu, and Yixuan Yuan.
\newblock Sigma: Semantic-complete graph matching for domain adaptive object detection.
\newblock In \emph{Proceedings of the IEEE/CVF conference on computer vision and pattern recognition}, pages 5291--5300, 2022{\natexlab{b}}.

\bibitem[Li et~al.(2023)Li, Guo, and Yuan]{li2023novel}
Wuyang Li, Xiaoqing Guo, and Yixuan Yuan.
\newblock Novel scenes \& classes: Towards adaptive open-set object detection.
\newblock In \emph{Proceedings of the IEEE/CVF international conference on computer vision}, pages 15780--15790, 2023.

\bibitem[Li et~al.(2021)Li, Chen, Xie, Yang, Yuan, Pu, and Zhuang]{li2021free}
Xianfeng Li, Weijie Chen, Di Xie, Shicai Yang, Peng Yuan, Shiliang Pu, and Yueting Zhuang.
\newblock A free lunch for unsupervised domain adaptive object detection without source data.
\newblock In \emph{Proceedings of the AAAI Conference on Artificial Intelligence}, pages 8474--8481, 2021.

\bibitem[Li and Liang(2021)]{li2021prefix}
Xiang~Lisa Li and Percy Liang.
\newblock Prefix-tuning: Optimizing continuous prompts for generation.
\newblock In \emph{Proceedings of the 59th Annual Meeting of the Association for Computational Linguistics and the 11th International Joint Conference on Natural Language Processing (Volume 1: Long Papers)}, pages 4582--4597, 2021.

\bibitem[Li et~al.(2018)Li, Wang, Shi, Hou, and Liu]{li2018adaptive}
Yanghao Li, Naiyan Wang, Jianping Shi, Xiaodi Hou, and Jiaying Liu.
\newblock Adaptive batch normalization for practical domain adaptation.
\newblock \emph{Pattern Recognition}, 80:\penalty0 109--117, 2018.

\bibitem[Liang et~al.(2023)Liang, Xue, Liu, Zhong, and Ming]{liang2023unknown}
Wenteng Liang, Feng Xue, Yihao Liu, Guofeng Zhong, and Anlong Ming.
\newblock Unknown sniffer for object detection: Don't turn a blind eye to unknown objects.
\newblock In \emph{Proceedings of the IEEE/CVF conference on computer vision and pattern recognition}, pages 3230--3239, 2023.

\bibitem[Lin et~al.(2017)Lin, Goyal, Girshick, He, and Doll{\'a}r]{lin2017focal}
Tsung-Yi Lin, Priya Goyal, Ross Girshick, Kaiming He, and Piotr Doll{\'a}r.
\newblock Focal loss for dense object detection.
\newblock In \emph{Proceedings of the IEEE international conference on computer vision}, pages 2980--2988, 2017.

\bibitem[Liu et~al.(2023)Liu, Lin, Shen, and Yang]{liu2023periodically}
Qipeng Liu, Luojun Lin, Zhifeng Shen, and Zhifeng Yang.
\newblock Periodically exchange teacher-student for source-free object detection.
\newblock In \emph{Proceedings of the IEEE/CVF international conference on computer vision}, pages 6414--6424, 2023.

\bibitem[Rezatofighi et~al.(2019)Rezatofighi, Tsoi, Gwak, Sadeghian, Reid, and Savarese]{rezatofighi2019generalized}
Hamid Rezatofighi, Nathan Tsoi, JunYoung Gwak, Amir Sadeghian, Ian Reid, and Silvio Savarese.
\newblock Generalized intersection over union: A metric and a loss for bounding box regression.
\newblock In \emph{Proceedings of the IEEE/CVF conference on computer vision and pattern recognition}, pages 658--666, 2019.

\bibitem[Saito et~al.(2018)Saito, Yamamoto, Ushiku, and Harada]{saito2018open}
Kuniaki Saito, Shohei Yamamoto, Yoshitaka Ushiku, and Tatsuya Harada.
\newblock Open set domain adaptation by backpropagation.
\newblock In \emph{Proceedings of the European conference on computer vision (ECCV)}, pages 153--168, 2018.

\bibitem[Sakaridis et~al.(2018)Sakaridis, Dai, and Van~Gool]{sakaridis2018semantic}
Christos Sakaridis, Dengxin Dai, and Luc Van~Gool.
\newblock Semantic foggy scene understanding with synthetic data.
\newblock \emph{International Journal of Computer Vision}, 126:\penalty0 973--992, 2018.

\bibitem[Sun et~al.(2024)Sun, Li, and Mu]{sun2024exploring}
Zhicheng Sun, Jinghan Li, and Yadong Mu.
\newblock Exploring orthogonality in open world object detection.
\newblock In \emph{Proceedings of the IEEE/CVF Conference on Computer Vision and Pattern Recognition}, pages 17302--17312, 2024.

\bibitem[Tarvainen and Valpola(2017)]{tarvainen2017mean}
Antti Tarvainen and Harri Valpola.
\newblock Mean teachers are better role models: Weight-averaged consistency targets improve semi-supervised deep learning results.
\newblock \emph{Advances in neural information processing systems}, 30, 2017.

\bibitem[Vs et~al.(2021)Vs, Gupta, Oza, Sindagi, and Patel]{vs2021mega}
Vibashan Vs, Vikram Gupta, Poojan Oza, Vishwanath~A Sindagi, and Vishal~M Patel.
\newblock Mega-cda: Memory guided attention for category-aware unsupervised domain adaptive object detection.
\newblock In \emph{Proceedings of the IEEE/CVF Conference on Computer Vision and Pattern Recognition}, pages 4516--4526, 2021.

\bibitem[VS et~al.(2023)VS, Oza, and Patel]{vs2023instance}
Vibashan VS, Poojan Oza, and Vishal~M Patel.
\newblock Instance relation graph guided source-free domain adaptive object detection.
\newblock In \emph{Proceedings of the IEEE/CVF conference on computer vision and pattern recognition}, pages 3520--3530, 2023.

\bibitem[Yu et~al.(2020)Yu, Chen, Wang, Xian, Chen, Liu, Madhavan, and Darrell]{yu2020bdd100k}
Fisher Yu, Haofeng Chen, Xin Wang, Wenqi Xian, Yingying Chen, Fangchen Liu, Vashisht Madhavan, and Trevor Darrell.
\newblock Bdd100k: A diverse driving dataset for heterogeneous multitask learning.
\newblock In \emph{Proceedings of the IEEE/CVF conference on computer vision and pattern recognition}, pages 2636--2645, 2020.

\bibitem[Zhu et~al.()Zhu, Su, Lu, Li, Wang, and Dai]{zhudeformable}
Xizhou Zhu, Weijie Su, Lewei Lu, Bin Li, Xiaogang Wang, and Jifeng Dai.
\newblock Deformable detr: Deformable transformers for end-to-end object detection.
\newblock In \emph{International Conference on Learning Representations}.

\bibitem[Zohar et~al.(2023)Zohar, Wang, and Yeung]{zohar2023prob}
Orr Zohar, Kuan-Chieh Wang, and Serena Yeung.
\newblock Prob: Probabilistic objectness for open world object detection.
\newblock In \emph{Proceedings of the IEEE/CVF Conference on Computer Vision and Pattern Recognition}, pages 11444--11453, 2023.

\end{thebibliography}
}

\clearpage
\setcounter{page}{1}
\setcounter{table}{0}
\setcounter{figure}{0}
\setcounter{equation}{0}

\captionsetup[table]{name=S.Table}
\captionsetup[figure]{name=S.Figure}

\maketitlesupplementary

\appendix

\section{Further Ablation Study}
\begin{figure}[t]
    \centering
    \includegraphics[width=\columnwidth]{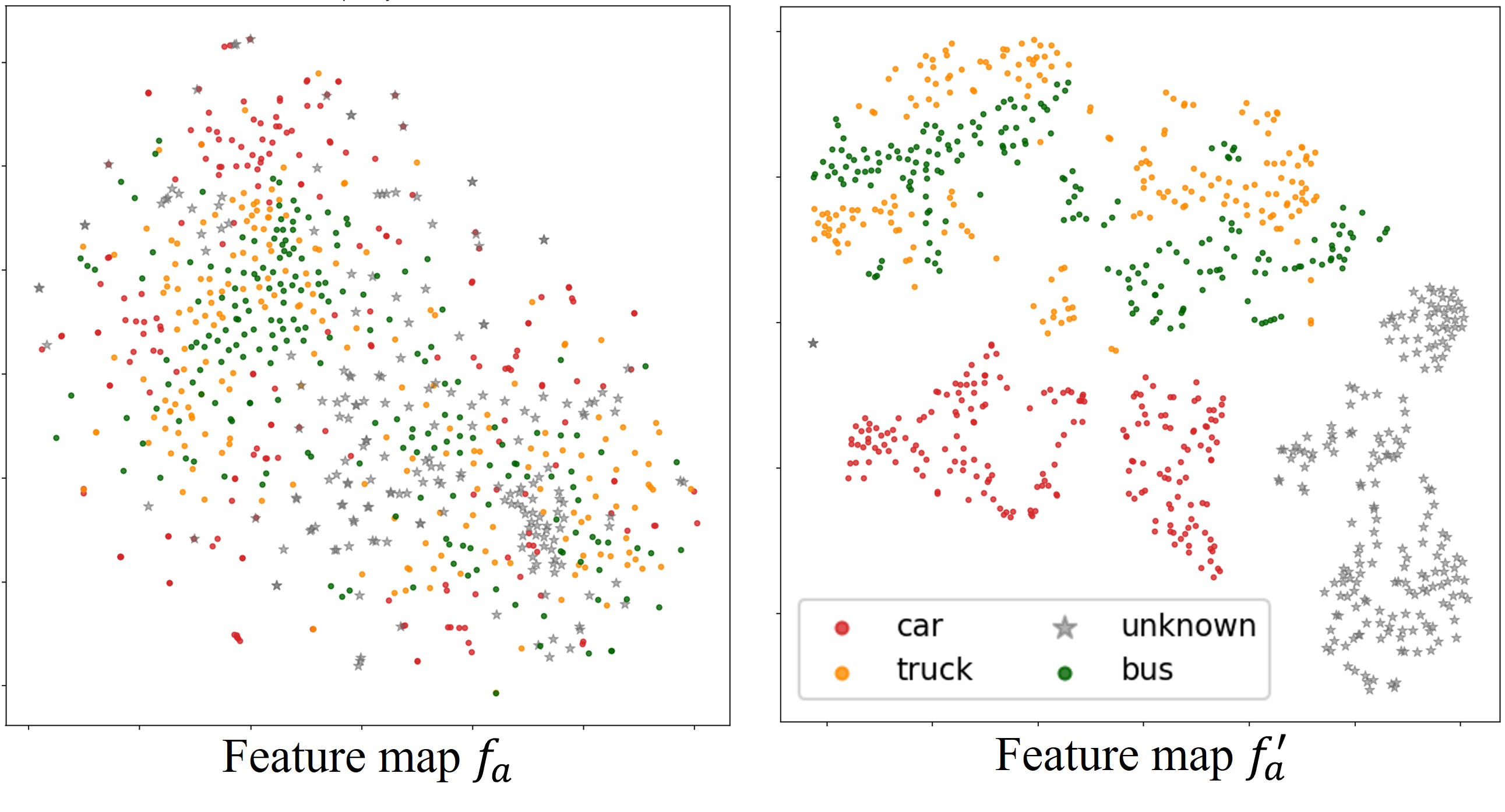}
    \caption{Visualization of the feature map.}
    \label{fig:viz_embeds}
\end{figure}

\paragraph{Ablation Study of the Target-dependent Knowledge.} Given the feature maps $f$ from the backbone, we select the top-$k$ activated features $f_a$ and apply truncated reconstruction with top-$r$ decomposed components via SVD to extract latent target knowledge shown in~S.Figure\ref{fig:viz_embeds}. To evaluate the effectiveness of top-k selection and top-r decomposition, we conducted ablation studies. S.Table~\ref{tab:ablation_recon_k} presents the results for selecting top-$k$ features from the feature maps. K=0 indicates truncated decomposition applied to all feature maps without selection. We found that $K$=50 achieved the best performance across known mAP, U-Recall, and H-Score. S.Table~\ref{tab:ablation_recon_r} compares performance variations with different top-$r$ values for truncated decomposition. Using 5 components (r=5) for the truncated reconstruction achieved a promising performance across all the metrics.

\paragraph{Ablation Study on the Threshold.} 
We further analyzed the effectiveness of the unknown threshold in generating the confidence mask for principal axis-based unknown labeling. As shown in S.Table~\ref{tab:ablation_unk_thr}, we evaluated the threshold $\epsilon$ within the range 0.1 to 0.9. Our findings indicate that setting $\epsilon$ = 0.3 achieves strong performance in U-Recall and H-Score. Notably, we also observed that increasing $\epsilon$ tends to improve known mAP but leads to a decline in U-Recall.

\begin{table}
\resizebox{\columnwidth}{!}{%
\setlength{\tabcolsep}{10pt} 
\renewcommand{\arraystretch}{1.} 
\begin{tabular}{c|cgy}
\specialrule{1pt}{0pt}{0pt}
\multirow{2}{*}{top-$k$ selection} & \multicolumn{3}{c}{Cityscapes $\rightarrow$ Foggy Cityscapes} \\ \cline{2-4}
 & Known mAP & U-Recall & H-Score \\ \hline
Baseline & 22.97 & 3.60 & 6.22 \\ \hdashline
$k=$ 0 & 29.07 & 7.52 & 11.95 \\
$k=$ 25 & \underline{31.84} & 7.23 & 11.78 \\ \hdashline
$\mathbf{k=}$ \textbf{50} & \textbf{32.32} & \textbf{10.59} & \textbf{15.95} \\ \hdashline
$k=$ 100 & 27.95 & \underline{8.27} & \underline{12.76} \\ 
\specialrule{1pt}{0pt}{0pt}
\end{tabular}%
}
\caption{Ablation study of selecting top-$k$ activated features.}
\label{tab:ablation_recon_k}
\vfill
\resizebox{\columnwidth}{!}{%
\setlength{\tabcolsep}{10pt} 
\renewcommand{\arraystretch}{1.} 
\begin{tabular}{c|cgy}
\specialrule{1pt}{0pt}{0pt}
\multirow{2}{*}{top-$r$ Recon} & \multicolumn{3}{c}{Cityscapes $\rightarrow$ Foggy Cityscapes} \\ \cline{2-4}
 & Known mAP & U-Recall & H-Score \\ \hline
$\mathbf{r=}$ \textbf{5} & \textbf{32.32} & \textbf{10.59} & \textbf{15.95} \\ \hdashline
$r=$ 10 & 30.77 & \underline{8.47} & \underline{13.28} \\
$r=$ 20 & 30.78 & 8.09 & 12.81 \\
$r=$ 30 & \underline{31.63} & 7.72 & 12.41 \\
\specialrule{1pt}{0pt}{0pt}
\end{tabular}%
}
\caption{Ablation study of the reconstruction on top-$r$}
\label{tab:ablation_recon_r}
\end{table}

\begin{table}
\resizebox{\columnwidth}{!}{%
\setlength{\tabcolsep}{10pt} 
\renewcommand{\arraystretch}{1.} 
\begin{tabular}{c|cgy}
\specialrule{1pt}{0pt}{0pt}
\multirow{2}{*}{Threshold $\epsilon$} & \multicolumn{3}{c}{Cityscapes $\rightarrow$ Foggy Cityscapes} \\ \cline{2-4}
 & Known mAP & U-Recall & H-Score \\ \hline
$\epsilon=$ 0.1 & 29.51 & 7.88 & 12.44 \\
$\boldsymbol{\epsilon=}$ \textbf{0.3} & 32.32 & \textbf{10.59} & \textbf{15.95} \\
$\epsilon=$ 0.5 & 32.81 & \underline{9.23} & \underline{14.41} \\
$\epsilon=$ 0.7 & \underline{33.45} & 7.36 & 12.07 \\
$\epsilon=$ 0.9 & \textbf{34.55} & 7.11 & 11.79 \\
\specialrule{1pt}{0pt}{0pt}
\end{tabular}%
}
\caption{Ablation study of the Unknown threshold $\epsilon$}
\label{tab:ablation_unk_thr}
\end{table}

\begin{table}
\resizebox{\columnwidth}{!}{%
\setlength{\tabcolsep}{10pt} 
\renewcommand{\arraystretch}{1.2} 
\begin{tabular}{c|cgy}
\specialrule{1pt}{0pt}{0pt}
\multirow{2}{*}{Unknown Pseudo Label} & \multicolumn{3}{c}{Cityscapes $\rightarrow$ Foggy Cityscapes} \\ \cline{2-4}
 & Known mAP & U-Recall & H-Score \\ \hline
OpenDet~\cite{han2022expanding} & \underline{42.09} & 1.79 & 3.43 \\
OW-DETR~\cite{gupta2022ow} & 39.92 & 1.98 & 3.77 \\
SOMA~\cite{li2023novel} & \textbf{45.55} & \underline{4.08} & \underline{7.49} \\ \hdashline
\textbf{CollaPAUL (Ours)} & 32.32 & \textbf{10.59} & \textbf{15.95} \\
\specialrule{1pt}{0pt}{0pt}
\end{tabular}%
}
\caption{Comparison of the performance on open-set methods.}
\label{tab:analysis_openset}
\end{table}

\section{Further Analysis}
\paragraph{Comparison with other Open-set Methods.}
To validate the effectiveness of the proposed CollaPAUL in detecting unknown objects as well as identifying known objects, we compared CollaPAUL with other open recognition methods: OpenDet\cite{han2022expanding}, OW-DETR\cite{gupta2022ow}, and SOMA~\cite{li2023novel}. OpenDet is designed to detect unknown objects during the training of annotated known objects. OW-DETR continuously learns annotated known objects and detects unannotated objects as unknowns. SOMA transfers knowledge for domain adaptation, enabling the detection of unknown objects while conveying knowledge of known objects from the source to the target domain.
However, these methods struggle to be applied to the proposed SFUOD setting because OpenDet assumes supervised learning, OW-DETR requires continuous learning in a single domain, and SOMA, which assumes domain-adaptive learning, depends on the source domain dataset for knowledge transfer.

Thus, we compared our performance at the proposed SFUOD scenario with these methods at the Adaptive Open-Set Object Detection (AOOD) scenario, where the detector can access both the labeled source dataset and the unlabeled target dataset during training. As shown in S.Table~\ref{tab:analysis_openset}, CollaPAUL achieved promising performance in both U-Recall and H-Score. Notably, unlike other methods that use labeled source domain datasets during training, CollaPAUL, which relies solely on unlabeled target domain datasets, achieved comparable performance in known mAP. These results demonstrate that CollaPAUL effectively detects unknown objects while maintaining strong performance compared to existing methods.

\end{document}